\documentclass[11pt]{article}
\usepackage{acl2013}
\usepackage{times}
\usepackage{url}
\usepackage{latexsym}
\usepackage{graphicx}
\usepackage{color}
\usepackage[skip=0pt]{caption}
\usepackage{subcaption}
\usepackage{subfig}
\usepackage{pgfplotstable}
\usepackage{adjustbox}
\usepackage{booktabs}
\usepackage{amssymb}
\usepackage{amsmath}
\usepackage{enumerate}
\usepackage{paralist}
\usepackage{xspace}

\newcommand{\RQ}[1][1]{\textbf{RQ#1}\xspace}

\newcommand{\figref}[2][]{Fig#1.~\ref{#2}\xspace}
\newcommand{\tabref}[2][]{Tab#1.~\ref{#2}\xspace}

\newcommand{\lex}[1]{\textit{#1}\xspace}

\newcommand{\dataset}[1]{\texttt{#1}\xspace}
\newcommand{\EWT}{\dataset{EWT}}
\newcommand{\WSJ}{\dataset{WSJ}}
\newcommand{\Brown}{\dataset{Brown}}
\newcommand{\Reuters}{\dataset{Reuters}}
\newcommand{\MUC}{\dataset{MUC7}}

\newcommand{\method}[2][]{\ensuremath{\textsc{#2#1}}\xspace}
\newcommand{\unigram}[1][]{\method{Unigram}}
\newcommand{\brown}[1][]{\method[\ensuremath{_{#1}}]{Brown}}
\newcommand{\CW}[1][]{\method[#1]{CW}}
\newcommand{\CBOW}[1][]{\method[#1]{CBOW}}
\newcommand{\Skipgram}[1][]{\method[#1]{Skip-gram}}
\newcommand{\Glove}[1][]{\method[#1]{Glove}}
\newcommand{\withup}{\method{+UP}}

\newcommand{\task}[1]{\textsf{#1}\xspace}
\newcommand{\pos}{\task{POS-tagging}}
\newcommand{\chunking}{\task{Chunking}}
\newcommand{\ner}{\task{NER}}
\newcommand{\mwe}{\task{MWE}}

\newcommand{\evmeasure}[1]{\textsc{#1}\xspace}
\newcommand{\accuracy}{\evmeasure{Acc}}
\newcommand{\fscore}{\evmeasure{F1}}

\newcommand{\best}[1]{\textbf{#1}}

\newcommand{\ctx}{\ensuremath{\text{ctx}}}

\title{Big Data Small Data, In Domain Out-of Domain, Known Word Unknown
  Word: The Impact of Word Representation on Sequence Labelling Tasks}
  
\author{
Lizhen Qu$^{1,2}$, Gabriela Ferraro$^{1,2}$, Liyuan Zhou$^1$, Weiwei Hou$^1$, Nathan Schneider$^3$ and Timothy Baldwin$^4$ \\
$^1$ NICTA, ACT 2601, Australia \\
$^2$ The Australian National University\\
$^3$ The University of Melbourne, VIC 3010, Australia \\
$^4$ University of Edinburgh, EH8 9AB, UK.  \\
{\tt \{lizhen.qu,gabriela.ferraro,liyuan.zho,weiwei.hou\}@nicta.com.au}\\
{\tt nschneid@cs.cmu.edu}\\
{\tt tb@ldwin.net}\\ 
~~\\
}

\date{}

\begin{document}

\maketitle

\begin{abstract} 
  Word embeddings -- distributed word representations that can be
  learned from unlabelled data -- have been shown to have high utility
  in many natural language processing applications. 
  In this paper, we perform an extrinsic evaluation of five popular word
  embedding methods in the context of four sequence labelling tasks:
  POS-tagging, syntactic chunking, NER and MWE identification.
  A particular focus of the paper is analysing the effects of task-based
  updating of word representations.
  We show that when using word embeddings as features, as few as
  several hundred training instances are sufficient to achieve competitive
  results, and that word embeddings lead to improvements over OOV words
  and out of domain.
  Perhaps more surprisingly, our results indicate there is little
  difference between the different word embedding methods, and that simple
  Brown clusters are often competitive with word embeddings across all
  tasks we consider. 
\end{abstract}

\newcommand{\gabi}[1]{\textcolor{blue}{#1}}
\newcommand{\tim}[1]{\textcolor{red}{#1}}
\newcommand{\lizhen}[1]{\textcolor{green}{#1}}
\newcommand{\nss}[1]{\textcolor{magenta}{#1}}

\section{Introduction}

Recently, distributed word representations have grown to become a
mainstay of natural language processing (NLP), and been show to have
empirical utility in a myriad of tasks
\cite{Collobert2008,turian2010word,baroni:2014,Andreas:Klein:2014}.  The
underlying idea behind distributed word representations is simple: to
map each word $w$ in our vocabulary $V$ onto a continuous-valued vector
of dimensionality $d \ll |V|$.  Words that are similar
(e.g., with respect to syntax or lexical semantics) will ideally be mapped to
similar regions of the vector space, implicitly supporting both
generalisation across in-vocabulary (IV) items, and countering the
effects of data sparsity for low-frequency and out-of-vocabulary (OOV)
items.

Without some means of automatically deriving the vector representations
without reliance on labelled data, however, word embeddings would have
little practical utility. Fortunately, it has been shown that they can
be ``pre-trained'' from unlabelled text data using various algorithms 
to model the distributional hypothesis (i.e., that
words which occur in similar contexts tend to be semantically
similar). Pre-training methods have been refined considerably in recent
years, and scaled up to increasingly large corpora.

As with other machine learning methods, it is well known that the
quality of the pre-trained word embeddings depends heavily on factors
including parameter optimisation, the size of the training data, and the
fit with the target application. For example, \newcite{turian2010word}
showed that the optimal dimensionality for word embeddings is task-specific.  
One factor which has received relatively little attention in
NLP is the effect of ``updating'' the pre-trained word embeddings as
part of the task-specific training, based on self-taught
learning~\cite{raina2007self}.  Updating leads to word
representations that are task-specific, but often at the cost of
over-fitting low-frequency and OOV words.



In this paper, we perform an extensive evaluation of four recently proposed word embedding
approaches under fixed experimental conditions, applied to four sequence
labelling tasks: POS-tagging, full-text chunking, named entity
recognition (NER), and multiword expression (MWE) identification. 
Compared to previous empirical studies~\cite{collobert2011natural,turian2010word,pennington2014glove}, we fill their gaps by considering more word embedding approaches 
and evaluating them with more sequence labelling tasks. In addition, we explore the following research questions:
\begin{compactenum}[\bf RQ1:]
\item are these word embeddings better than baseline approaches of one-hot
  unigram features and Brown clusters?
\item do word embeddings require less training data (i.e.\ generalise
  better) than one-hot unigram features? If so, to what degree can word embeddings reduce the amount of labelled data?
\item what is the impact of updating word embeddings in sequence
  labelling tasks, both empirically over the target task and
  geometrically over the vectors?
\item what is the impact of these word embeddings (with and without
  updating) on both OOV items (relative to the training data) and
  out-of-domain data?
\item overall, are some word embeddings better than others in a sequence
  labelling context?
\end{compactenum}




\section{Word Representations}
\label{wordrep} 

\subsection{Types of Word Representations}

\newcite{turian2010word} identifies three varieties of word
representations: \textit{distributional}, \textit{cluster-based}, and
\textit{distributed}.

\textit{Distributional representation} methods map each word $w$ to a
context word vector $\mathbf{C}_w$, which is constructed directly from
co-occurrence counts between $w$ and its context words.  The learning
methods either store the co-occurrence counts between two words $w$ and
$i$ directly in
$C_{wi}$~\cite{sahlgren2006word,turney2010frequency,honkela1997self} or
project the concurrence counts between words into a lower dimensional
space~\cite{vrehuuvrek2010software,lund1996producing}, using
dimensionality reduction techniques such as SVD~\cite{dumais1988using} and
LDA~\cite{blei2003latent}.

\textit{Cluster-based representation} methods build clusters of words by applying either soft or hard clustering algorithms~\cite{lin2009phrase,li2005semi}. Some of them also rely on a co-occurrence matrix of words~\cite{pereira1993distributional}. The Brown clustering algorithm~\cite{Brown92class-basedn-gram} is the best-known method in this category.

\textit{Distributed representation} methods usually map words into dense,
low-dimensional, continuous-valued vectors, with $\mathbf{x} \in
R^d$, where $d$ is referred to as the word dimension.

\subsection{Selected Word Representations}

Over a range of sequence labelling tasks, we evaluate five methods for
inducing word representations: Brown clustering
\cite{Brown92class-basedn-gram} (``\brown''), the neural language model
of Collobert \& Weston (``\CW'')~\cite{collobert2011natural}, the
continuous bag-of-words model (``\CBOW'')~\cite{Mikolov13}, the continuous
skip-gram model (``\Skipgram'')~\cite{Mikolov13NIPS}, and Global vectors
(``\Glove'')~\cite{pennington2014glove}. With the exception of \CW, all have
have been shown to be at or near state-of-the-art in recent empirical
studies~\cite{turian2010word,pennington2014glove}. \CW is included
because it was highly influential in earlier research, and the pre-trained embeddings are
still used to some degree in NLP. The training of these word
representations is unsupervised: the common underlying idea is to
predict occurrence of words in the neighbouring context. Their training
objectives share the same form, which is a sum of local training factors
$J(w, \text{ctx}(w))$,  
\begin{displaymath}
  L = \sum_{w \in V} J(w, \ctx(w))
\end{displaymath}

where $V$ is the vocabulary of a given corpus, and $\ctx(w)$ denotes the
local context of word $w$.
The local context of a word can either be its previous $k$ words, or the
$k$ words surrounding it. 
Local training factors are designed to capture the relationship between
$w$ and its local contexts of use, either by predicting $w$
based on its local context, or using $w$ to predict the
context words. Other than \brown, which utilises a cluster-based
representation, all the other methods employ a distributed representation.

The starting point for \CBOW and \Skipgram is to employ softmax to predict word occurrence:
\begin{displaymath}
  J(w, \ctx(w)) = - \log \left( \frac{\exp(\mathbf{v}_w^{\text{T}} \mathbf{v}_{\ctx(w)})}{ \sum_{j \in V} \exp(\mathbf{v}_j^{\text{T}} \mathbf{v}_{\ctx(w)})} \right)
\end{displaymath}
where $\mathbf{v}_{\ctx(w)}$ denotes the distributed representation of
the local context of word $w$. \CBOW derives $\mathbf{v}_{\ctx(w)}$
based on averaging over the context words. That is, it estimates the
probability of each $w$ given its local
context. In contrast, \Skipgram applies softmax to each context word of
a given occurrence of word $w$. In this case, $\mathbf{v}_{\ctx(w)}$ corresponds to the
representation of one of its context words. This model can be characterised as
predicting context words based on $w$. In practice, softmax is
too expensive to compute over large corpora, and thus~\newcite{Mikolov13NIPS} use
hierarchical softmax and negative sampling to scale up the training.

\CW considers the local context of a word $w$ to be $m$ words to the left
and $m$ words to the right of $w$. The concatenation of the embeddings of
$w$ and all its context words are taken as input to a neural network
with one hidden layer, which produces a higher level representation
$f(w) \in R^d$. Then the learning procedure replaces the embedding of
$w$ with that of a randomly sampled word $w'$ and generates a second
representation $f(w') \in R^d$ with the same neural network. The
training objective is to maximise the difference between them:
\begin{displaymath}
J(w, \ctx(w)) = \max (0, 1 - f(w) + f(w'))
\end{displaymath}
This approach can be regarded as negative sampling with only one negative example.

\Glove assumes the dot product of two word embeddings should be similar
to the logarithm of the co-occurrence count $X_{ij}$ of the two
words. As such, the local factor $J(w, \ctx(w))$ becomes:
\begin{displaymath}
g(X_{ij}) (\mathbf{v}_i^{\text{T}} \mathbf{v}_j + b_i + b_j - \log(X_{ij}))^2
\end{displaymath}
where $b_i$ and $b_j$ are the bias terms of words $i$ and $j$,
respectively, and $g(X_{ij})$ is a weighting function based on the
co-occurrence count. This weighting function controls the degree of
agreement between the parametric function $\mathbf{v}_i^{\text{T}}
\mathbf{v}_j + b_i + b_j $ and $\log(X_{ij})$. Frequently co-occurring
word pairs will be larger weight
than infrequent
pairs, up to a threshold.

\brown partitions words into a finite set of word classes $V$. The
conditional probability of seeing the next word is defined to be:
\begin{displaymath}
p(w_k | w_{k - m}^{k -1}) = p(w_k | h_k) p(h_k | h_{k - m}^{k -1})
\end{displaymath}
where $h_k$ denotes the word class of the word $w_k$, $w_{k - m}^{k -1}$
are the previous $m$ words and $h_{k - m}^{k -1}$ are their respective
word classes. Then $J(w, \text{ctx}(w)) = - \log p(w_k | w_{k - m}^{k
  -1}) $. Since there is no tractable method to find an optimal
partition of word classes, the method uses only a bigram class model, and utilises hierarchical clustering as an approximation method to find a sufficiently good partition of words.

\subsection{Building Word Representations}
\label{buildingWordRep}

For a fair comparison, we train \brown, \CBOW, \Skipgram, and \Glove on
a fixed corpus, comprised of freely available corpora, as detailed in
\tabref{wordEmbedCorpora}. The joint corpus was preprocessed with the
Stanford CoreNLP sentence splitter and tokeniser. All consecutive digit
substrings were replaced by NUM\textit{f}, where \textit{f} is the
length of the digit substring (e.g., \lex{10.20} is replaced by
\lex{NUM2.NUM2}. Due to the computational complexity of the
pre-training, for \CW, we simply downloaded the pre-compiled embeddings from:
\url{http://metaoptimize.com/projects/wordreprs}.

\begin{table}[t]
\centering
\begin{tabular}{lrr}
\hline
\textbf{Data set} & \multicolumn{1}{c}{\textbf{Size}} & \multicolumn{1}{c}{\textbf{Words}} \\ \hline
UMBC 	& 48.1GB & 3G \\
One Billion 	& 4.1GB & 1G  \\
English Wikipedia & 49.6GB & 3G \\ \hline
\end{tabular}
\caption{Corpora used to pre-train the word embeddings}
\label{wordEmbedCorpora}
\end{table}

The dimensionality of the word embeddings and the size of the context
window are the key hyperparameters when learning distributed
representations. We use all combinations of the following values to
train word embeddings on the combined corpus:
\begin{itemize}
\item \textbf{Embedding dim.\ $d \in \{25, 50, 100, 200\}$}
\item \textbf{Context window size $m \in \{1, 5, 10\}$}
\end{itemize}
\brown requires only the number of clusters as a hyperparameter. We
perform clustering with $b \in \{250, 500, 1000, 2000, 4000\}$ clusters.

\section{Sequence Labelling Tasks}
\label{sec:SeqTagging}

We evaluate the different word representations over four sequence
labelling tasks: POS-tagging (\pos), full-text chunking (\chunking),
NER (\ner) and MWE identification (\mwe). For
each task, we fed features into a first order linear-chain graph
transformer~\cite{collobert2011natural} made up of two layers: the upper
layer is identical to a linear-chain CRF~\cite{lafferty2001conditional},
and the lower layer consists of word representation and hand-crafted
features. If we treat word representations as fixed, the graph
transformer is a simple linear-chain CRF. On the other hand, if we can
treat the word representations as model parameters, the model is
equivalent to a neural network with word embeddings as the input
layer. We trained all models using AdaGrad~\cite{duchi2011adaptive}.


As in~\newcite{turian2010word}, at each word position, we construct word
representation features from the words in a context window of size two
to either side of the target word, based on the pre-trained
representation of each word type.  For \brown, the features are the
prefix features extracted from word clusters in the same way
as~\newcite{turian2010word}. As a baseline (and to test \RQ[1]), we include a one-hot
representation (which is equivalent to a linear-chain CRF with only
lexical context features).

Our hand-crafted features for \pos, \chunking and \mwe, are those used
by \newcite{collobert2011natural}, \newcite{turian2010word}
and~\newcite{mwecorpus}, respectively. For \ner, we use the same feature
space as \newcite{turian2010word}, except for the previous two
predictions, because we want to evaluate all word representations with
the same type of model -- a first-order graph transformer.

In training the distributed word representations, we consider two
settings: (1) the word representations are fixed during sequence model
training; and (2) the graph transformer updated the token-level word
representations during training.

\begin{table*}
\begin{small}
\begin{tabular}{@{}c@{~~}c@{~~}c@{~~}c@{~~}c@{~~}c@{}}
\hline
			& \textbf{Training} & \textbf{Development} & \textbf{\textit{In-domain} Test} & \textbf{\textit{Out-of-domain} Test} \\ \hline
\textbf{\pos} & \WSJ Sec.\ 0-18  & \WSJ Sec.\ 19--21 & \WSJ Sec.\ 22--24 & \EWT  \\
\textbf{\chunking} & \WSJ & \WSJ (1K sentences) & \WSJ (CoNLL-00 test) & \Brown \\
\textbf{\ner} & \Reuters (CoNLL-03 train) & \Reuters (CoNLL-03 dev) & \Reuters (CoNLL-03 test) & \MUC  \\
\textbf{\mwe} & \EWT (500 docs) & \EWT (100 docs)  & \EWT (123 docs) & --- \\
\hline
\end{tabular}
\caption{Training, development and test (in- and out-of-domain) data for each sequence labelling
  task.}
\label{datasplit}
\end{small}
\end{table*}

As outlined in \tabref{datasplit}, for each sequence labelling task, we
experiment over the de facto corpus, based on pre-existing
training--dev--test splits where available:\footnote{For the \mwe
  dataset, no such split pre-existed, so we constructed our own.}
\begin{compactenum}
\item[\textbf{\pos}:] the Wall Street Journal portion of the Penn
  Treebank (\newcite{Marcus:1993}: ``\WSJ'')
  with Penn POS tags
\item[\textbf{\chunking}:] the Wall Street Journal portion of the Penn
  Treebank (``\WSJ''),
  converted into IOB-style full-text chunks using the CoNLL conversion
  scripts for training and dev, and the WSJ-derived CoNLL-2000 full text chunking
  test data for testing \cite{TjongKimSang:Buchholz:2000}
\item[\textbf{\ner}:] the English portion of the CoNLL-2003 English Named Entity Recognition
  data set, for which the source data was taken from Reuters newswire
  articles (\newcite{TjongKimSang:DeMeulder:2003}: ``\Reuters'')
\item[\textbf{\mwe}:] the MWE dataset of \newcite{mwecorpus}, over a portion of text from the
  English Web Treebank\footnote{\url{https://catalog.ldc.upenn.edu/LDC2012T13}} (``\EWT'')
\end{compactenum}
 For all tasks other
than \mwe,\footnote{Unfortunately, there is no
  second domain which has been hand-tagged with MWEs using the method of
  \newcite{mwecorpus} to use as an out-of-domain test corpus.} we
additionally have an out-of-domain test set, in order to evaluate the
out-of-domain robustness of the different word representations, with and
without updating. These datasets are as follows:
\begin{compactenum}
\item[\textbf{\pos}:] the English Web Treebank with Penn POS tags (``\EWT'')
\item[\textbf{\chunking}:] the Brown Corpus portion of the Penn
  Treebank (``\Brown''), 
  converted into IOB-style full-text chunks using the CoNLL conversion
  scripts
\item[\textbf{\ner}:] the MUC-7 named entity recognition corpus\footnote{https://catalog.ldc.upenn.edu/LDC2001T02} (``\MUC'')
\end{compactenum}

For reproducibility, we tuned the hyperparameters with random search
over the development data for each task~\cite{bergstra2012random}. 
In this, we randomly sampled 50 distinct hyperparameter sets with the
same random seed for the non-updating models (i.e.\ the models that
don't update the word representation), and
sampled 100 distinct hyperparameter sets for the updating models (i.e.\
the models that do). 
For each set of hyperparameters and task, we train a model over its
training set and choose the best one based on its performance on development data~\cite{turian2010word}. 
We also tune the word representation hyperparameters -- namely, the word
vector size $d$ and context window size $m$ (distributed
representations), and in the case of \Brown, the number of clusters.

For the updating models, we found that the results over the test data
were always inferior to those that do not update the word
representations, due to the higher number of hyperparameters and small
sample size (i.e.\ 100).
Since the two-layer model of the graph transformer contains a distinct
set of hyperparameters for each layer, we reuse the best-performing
hyperparameter settings from the non-updating models, and only tune the
hyperparameters of AdaGrad for the word representation layer. 
This method requires only 32 additional runs and achieves consistently
better results than 100 random draws.

In order to test the impact of the volume of training data on the
different models (\RQ[2]), we split the training set into 10 partitions based on
a base-2 log scale (i.e., the second smallest partition will be twice
the size of the smallest partition), and created 10 successively larger
training sets by merging these partitions from the smallest one to the
largest one, and used each of these to train a model.
From these, we construct learning curves over each task.

For ease of comparison with previous results, we evaluate both
in- and out-of-domain using 
chunk/entity/expression-level F1-measure (``\fscore'') for all tasks except \pos,
for which we use token-level accuracy (``\accuracy'').
To test performance over OOV (unknown) tokens -- i.e., the words that do
not occur in the training set -- we use token-level accuracy for all
tasks (e.g.\ for \chunking, we evaluate whether the full IOB tag is
correct or not), due to the sparsity of all-OOV chunks/NEs/MWEs.






\section{Experimental Results and Discussion}

\begin{table*}[t]
\begin{center}
\begin{small}
\begin{tabular}{llll}
\hline
\textbf{Task}  & \textbf{Benchmark} & \textbf{\textit{In-domain} Test set} & \textbf{\textit{Out-of-domain} Test set} \\ \hline
\pos (\accuracy)    & \best{0.972} \cite{Toutanova:2003} & 0.959 (\Skipgram[\withup]) & 0.910 (\Skipgram)\\ 
\chunking (\fscore) & \best{0.942} \cite{Sha:2003} & 0.938 (\brown[b = 2000]) & 0.676 (\Glove)\\  
\ner (\fscore)      & \best{0.893} \cite{Ando:2005} & 0.868 (\Skipgram) & 0.736 (\Skipgram) \\  
\mwe (\fscore)      &0.625 \cite{Schneider+:2014} & \best{0.654} (\CBOW[\withup]) & --- \\ %
\hline
\end{tabular}
\caption{State-of-the-art results vs.\ our best results for in-domain and
  out-of-domain test sets.}
\label{benchmark}
\end{small}
\end{center}
\end{table*}

\begin{figure*}[t!]
\centering
\begin{subfigure}{7cm}
	\centering
    \includegraphics[scale=0.38]{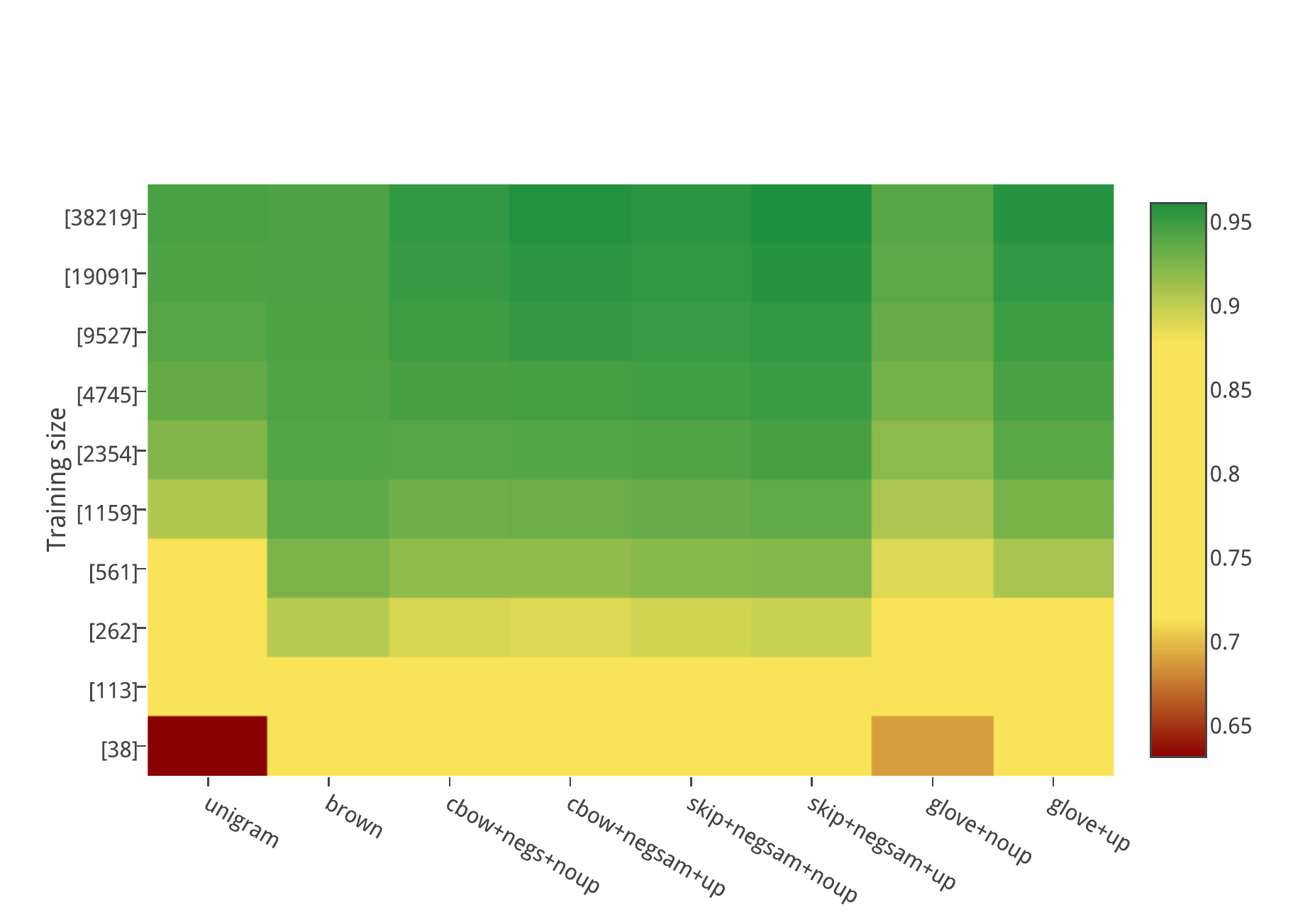}    	
	\subcaption{\pos (\accuracy)}	
	\label{pos}
\end{subfigure}
\begin{subfigure}{7cm}
	\centering
    \includegraphics[scale=0.38]{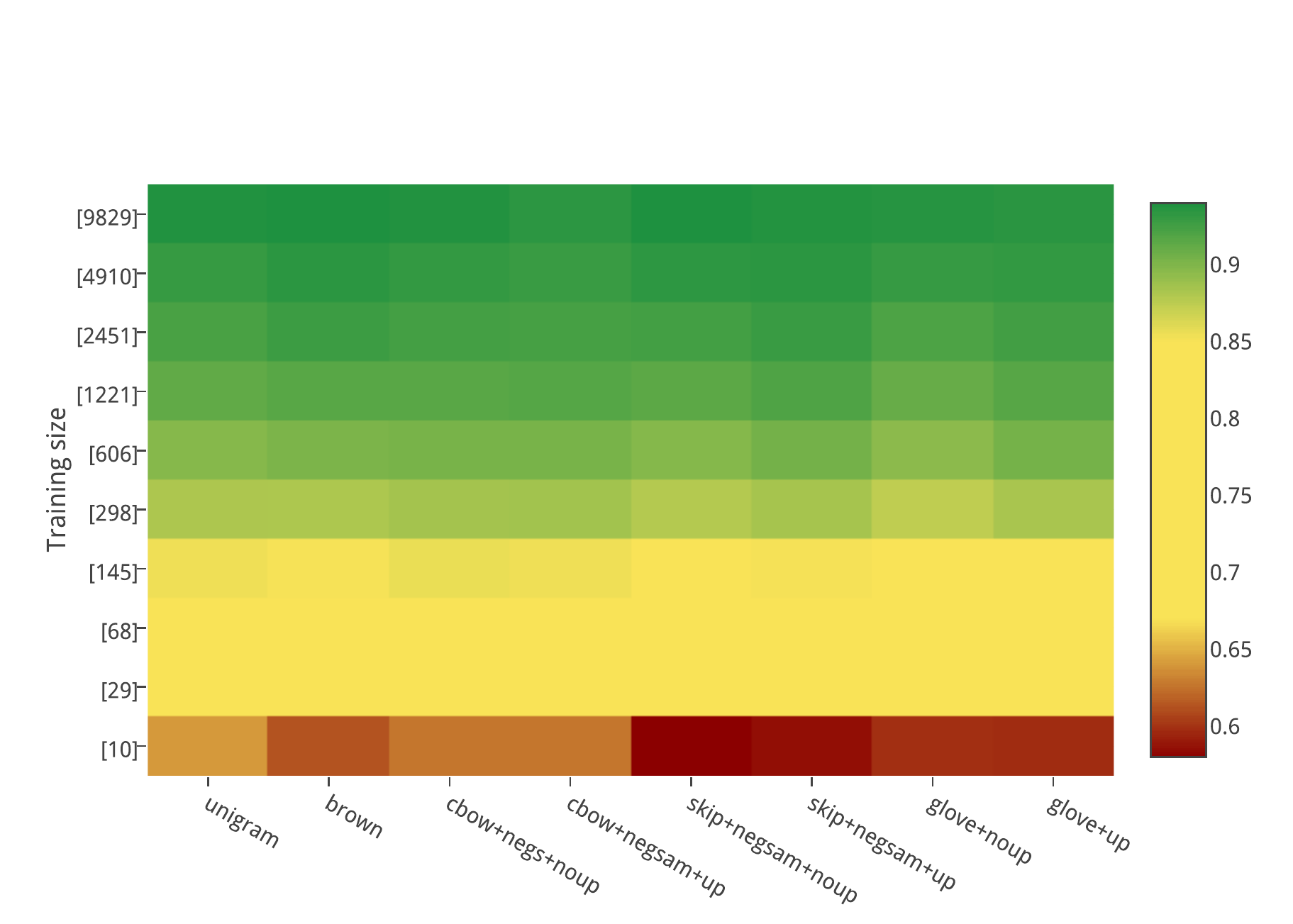}
	\subcaption{\chunking (\fscore)}	
	\label{chu}
\end{subfigure}
\begin{subfigure}{7cm}
	\centering
    \includegraphics[scale=0.38]{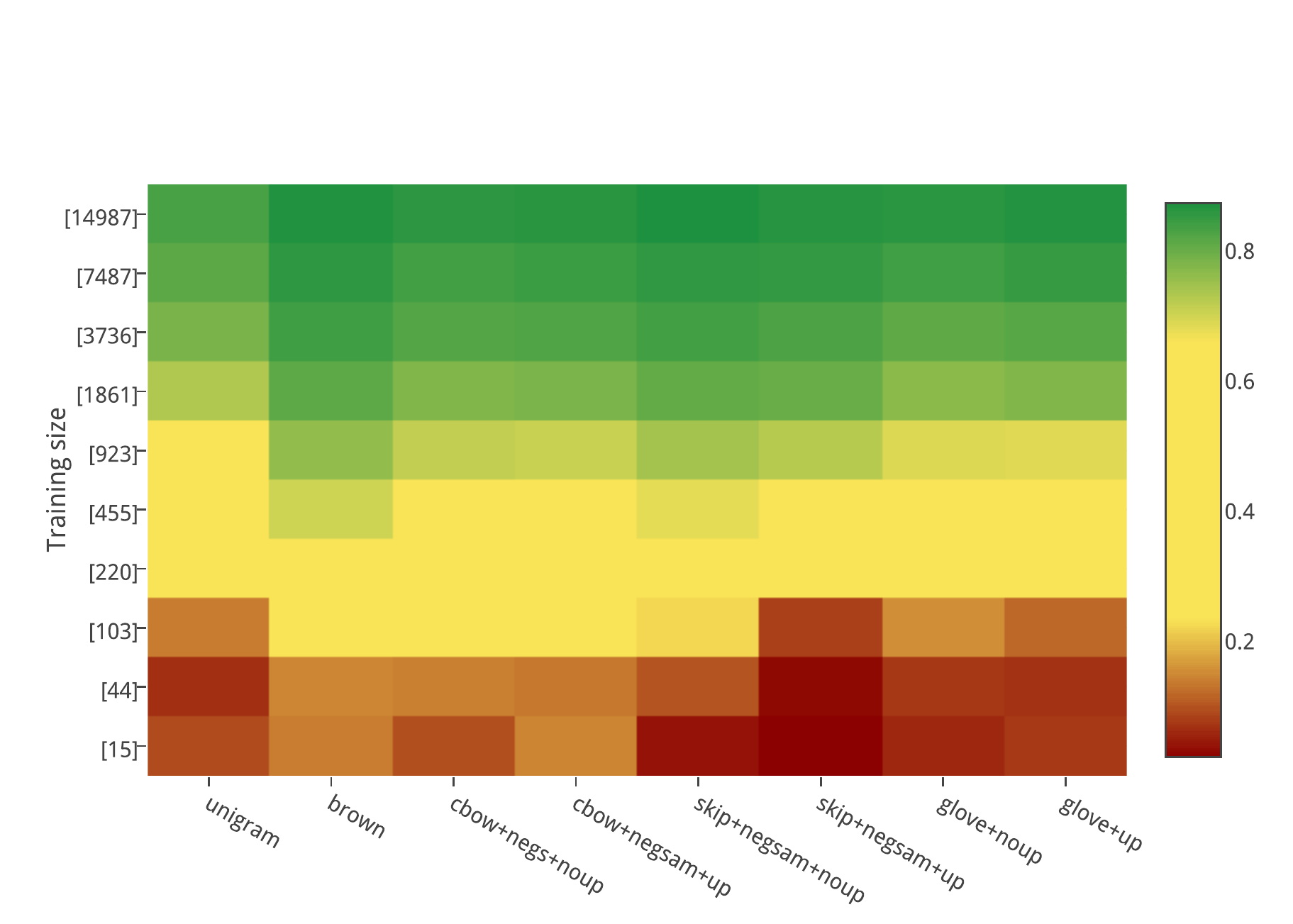}    	
	\subcaption{\ner (\fscore)}	
	\label{ner}
\end{subfigure}
\begin{subfigure}{7cm}
	\centering
    \includegraphics[scale=0.38]{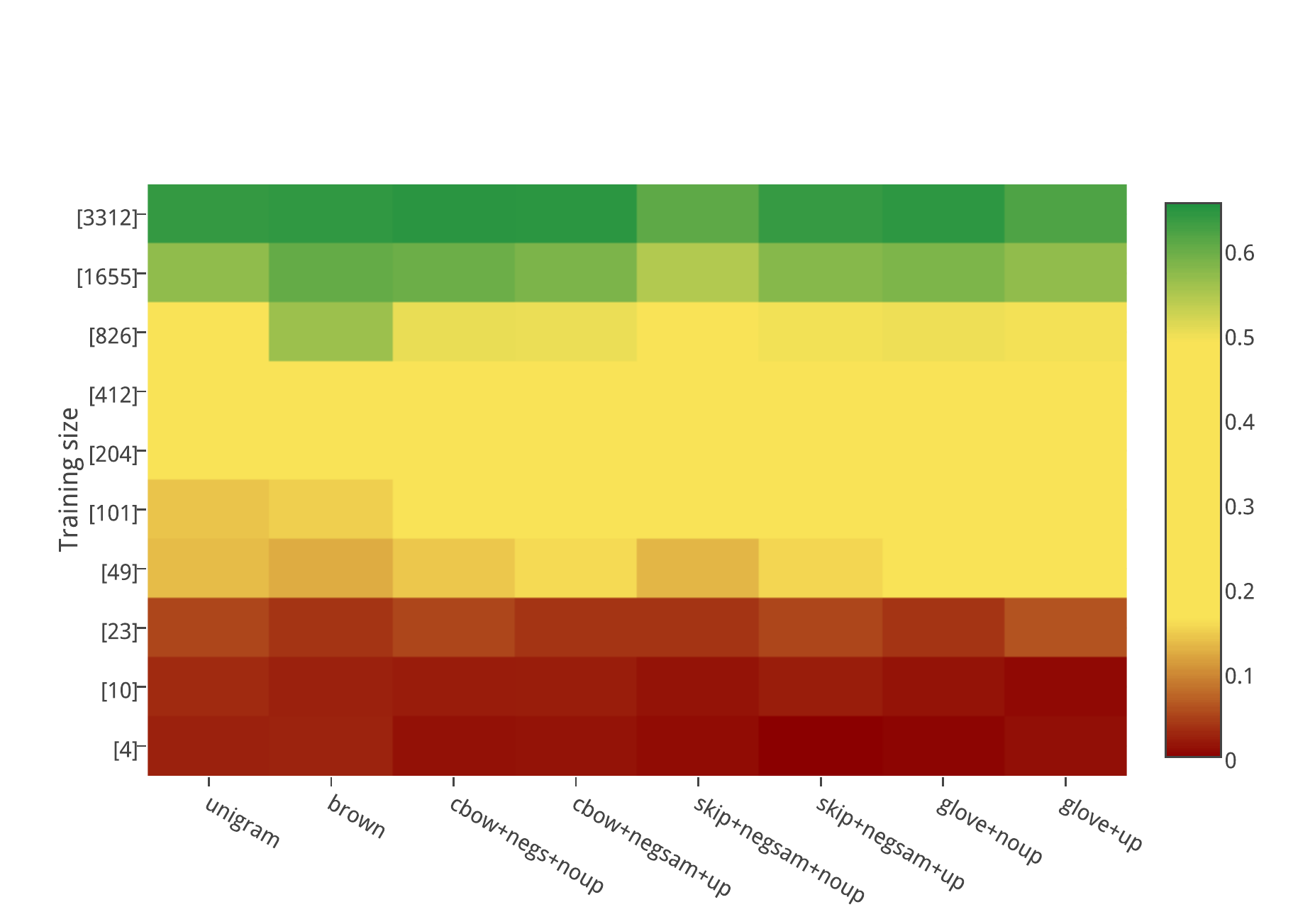}
	\subcaption{\mwe (\fscore)}		
	\label{mwe}
\end{subfigure}
\caption{Results for each type of word representation over \pos, \chunking, \ner and
  \mwe, optionally with updating (``\withup''). The $y$-axis indicates the training data
  sizes (on a log scale). Green = high performance, and red = low
  performance, based on a linear scale of the best- to worst-result for
  each task. }
\label{fig:heatmaps}
\end{figure*}

\begin{figure*}[t!]
\centering
    	\includegraphics[scale=0.5]{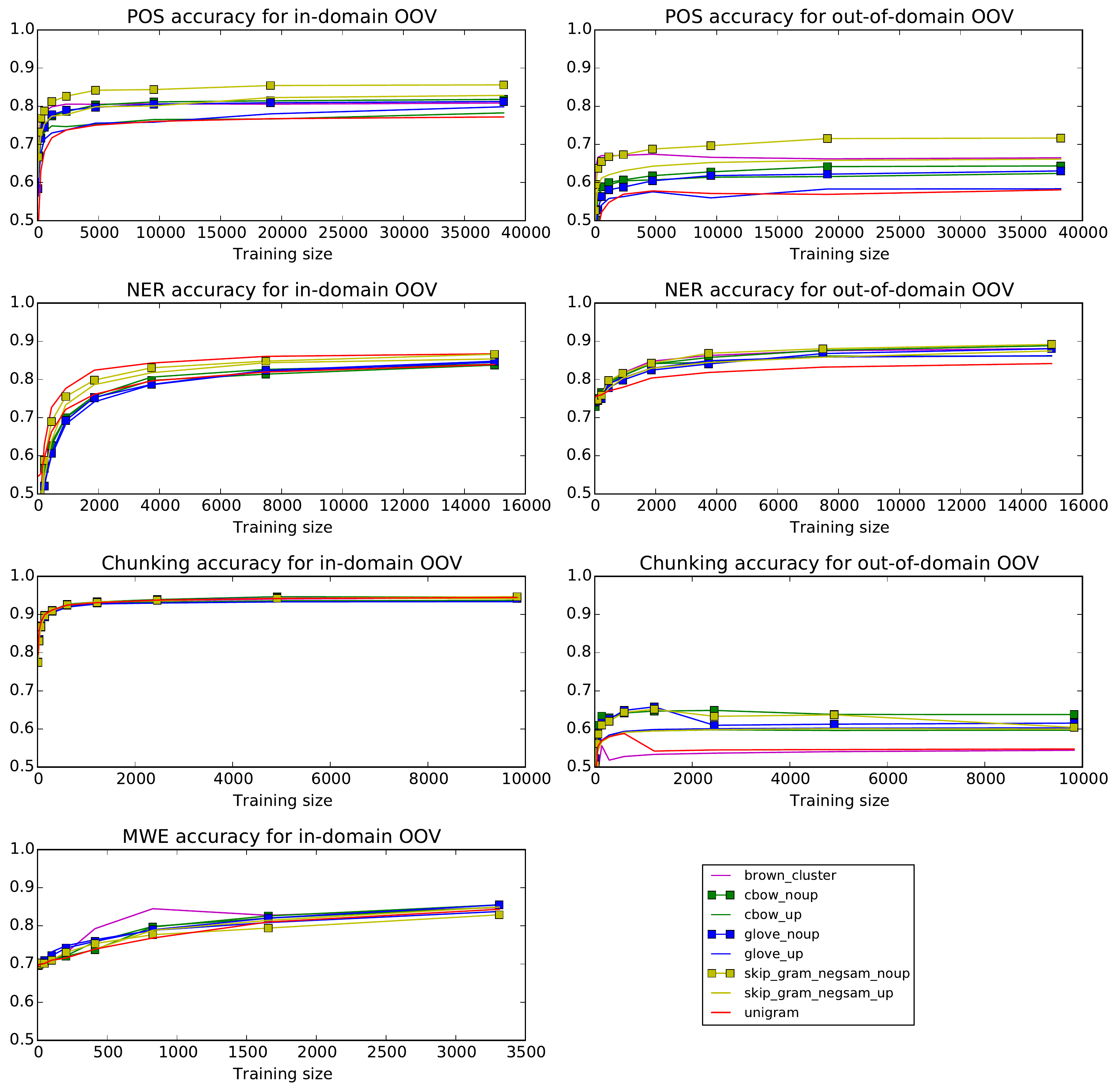}
\caption{\accuracy over out-of-vocabulary (OOV) words for \textit{in-domain} and \textit{out-of-domain} test sets.}
\label{OOV} 
\end{figure*}

\begin{figure*}[t!]
\centering
\begin{subfigure}[b]{0.48\textwidth}
	\centering
    \includegraphics[width=\textwidth]{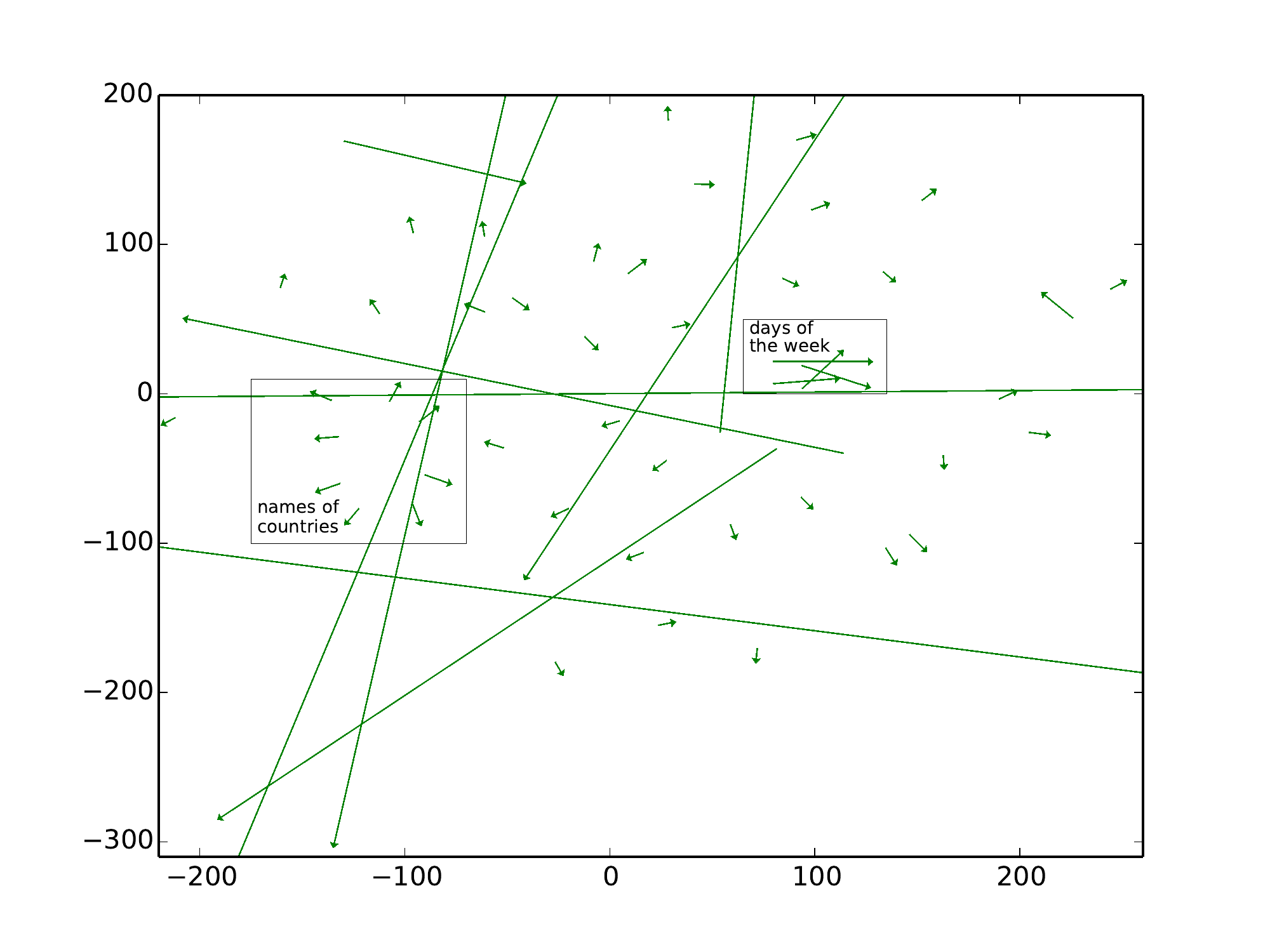}
	\subcaption{\chunking}	
	\label{fig:skipChu}
\end{subfigure}
\begin{subfigure}[b]{0.48\textwidth}
	\centering
    \includegraphics[width=\textwidth]{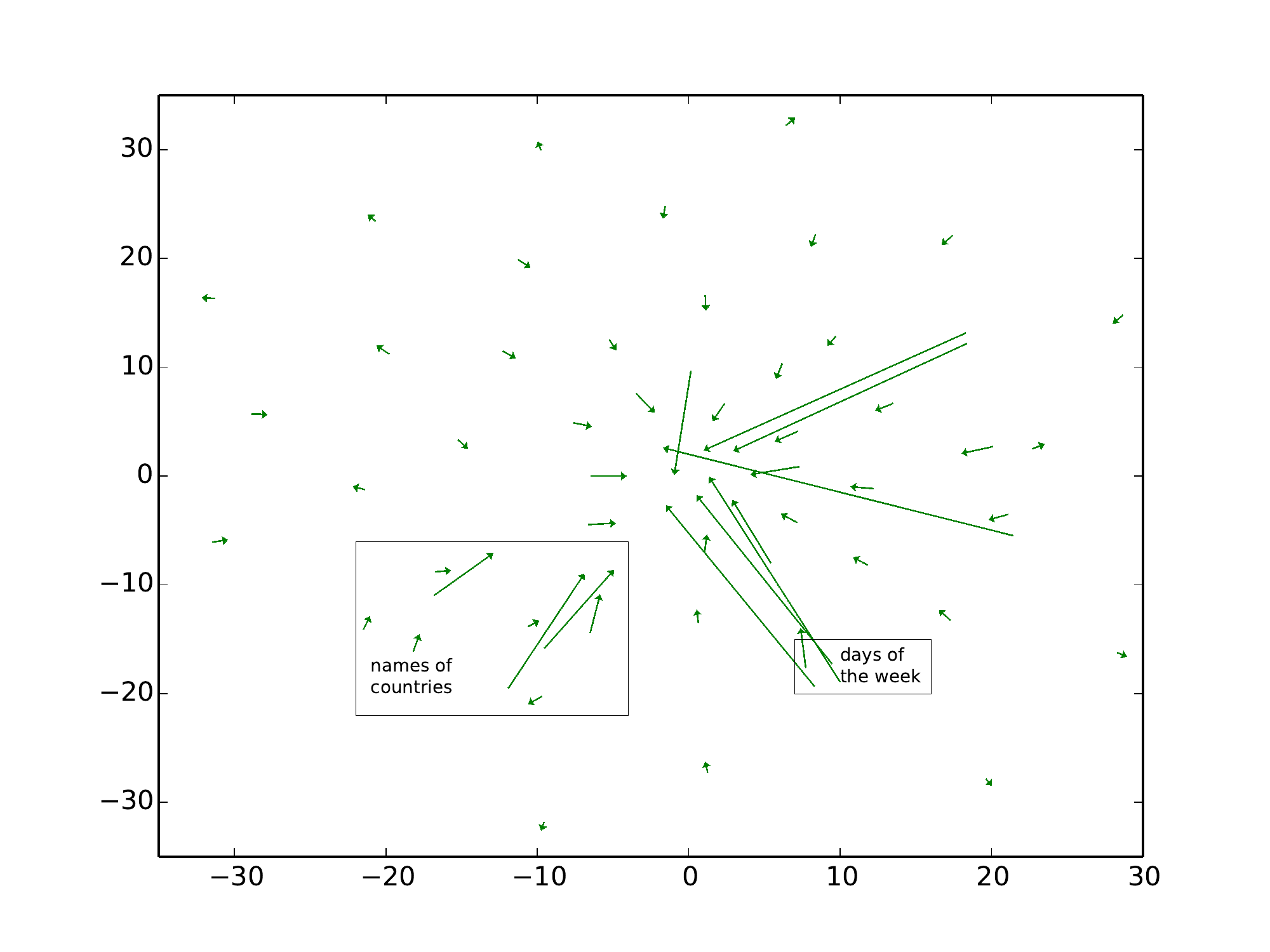}    	
	\subcaption{\ner}
	\label{fig:skippos}	
\end{subfigure}
\caption{A t-SNE plot of the impact of updating on \Skipgram}
\label{fig:vectorfield}
\end{figure*}

We structure our evaluation by stepping through each of our five
research questions (\RQ[1--5]) from the start of the paper. In this, we
make reference to: (1) the best-performing method both in- and
out-of-domain vs.\ the state-of-the-art (\tabref{benchmark}); (2) a
heat map for each task indicating the convergence rate for each word
representation, with and without updating (\figref{fig:heatmaps}); 
(3) OOV accuracy both in-domain and out-of-domain for each task
(\figref{OOV}); and (4)  visualisation of the impact of
updating on word embeddings, based on t-SNE
(\figref{fig:vectorfield}).

\paragraph{\RQ[1]: Are the selected word embeddings better than one-hot unigram features
  and Brown clusters?}  As shown in \tabref{benchmark}, the
best-performing method for every task except in-domain \chunking is a
word embedding method, although the precise method varies
greatly. \figref{fig:heatmaps}, on the other hand, tells a more subtle
story: the difference between \unigram and the other word
representations is relatively modest, esp.\ as the amount of training
data increases. Additionally, the difference between \brown and the word
embedding methods is modest across all tasks. So, the overall answer
would appear to be: yes for unigrams when there is little training data, but not really for \brown.

\paragraph{\RQ[2]: Do word embedding features require less training data?}
\figref{fig:heatmaps} shows that for \pos and \ner, with only several hundred training instances, 
word embedding features achieve superior results to \unigram. 
For example, when trained with 561 instances, the \pos model using \Skipgram[\withup] embeddings is 5.3\% above
\unigram; and when trained with 932 instances, the \ner model using \Skipgram is 11.7\% above \unigram. 
Similar improvements are also found for other types of word embeddings and \brown, when the training set is small. 
However, all word representations perform similarly for \chunking
regardless of training data size.
For \mwe, \brown performs slightly better than the other methods when
trained with approximately 25\% of the training instances. 
Therefore, we conjecture that the \pos and \ner tasks benefit more from
distributional similarity than \chunking and \mwe.

\paragraph{\RQ[3]: Does task-specific updating improve all word embeddings across all tasks?}
Based on \figref{fig:heatmaps}, updating of word representations can
equally correct poorly-learned word representations, and harm
pre-trained representations, due to overfitting.
For example, 
\Glove perform significantly worse than \Skipgram
in both \pos and \ner without updating, but \emph{with} updating, the
gap between their results and the best performing method becomes
smaller. In contrast, \Skipgram performs worse over the test data with
updating, despite the results on the development set improving by 1\%.

To further investigate the effects of updating, we sampled 60 words and
plotted the changes in their word embeddings under updating, using 2-d
vector fields generated by using matplotlib and t-SNE \cite{vanderMaaten:Hinton:2008}. Half
of the words were chosen manually to include known word clusters such as
days of the week and names of countries; the other half were selected
randomly. Additional plots with 100 randomly-sampled words and the
top-100 most frequent words, for all the methods and all the tasks, can
be found in the supplementary material and at
\url{https://123abc123abd.wordpress.com/}.  In each plot, a single arrow
signifies one word, pointing from the position of the original word embedding to the updated representation.

In \figref{fig:vectorfield}, we show vector fields plots for \chunking and \ner using \Skipgram embeddings.
For \chunking, most of the vectors were changed with similar magnitude,
but in very different directions, including within the clusters of days of
the week and country names.
In contrast, for \ner, there was more homogeneous change in word vectors
belonging to the same cluster.
This greater consistency is further evidence that semantic homogeneity
appears to be more beneficial for \ner than \chunking.

\paragraph{\RQ[4]: What is the impact of word embeddings cross-domain
  and for OOV words?}
As shown in \tabref{benchmark}, results predictably drop when we
evaluate out of domain.
The difference is most pronounced for \chunking, where there is an
absolute drop in \fscore of around 30\% for all methods, indicating that
word embeddings and unigram features provide similar information for
\chunking. 

Another interesting observation is that updating often hurts
out-of-domain performance because the distribution between domains is different. 
This suggests that, if the objective is to optimise performance across
domains, it is best not to perform updating. 

We also analyze performance on OOV words both in-domain and
out-of-domain in \figref{OOV}.
As expected, word embeddings and \brown excel in out-of-domain OOV performance.
Consistent with our overall observations about cross-domain
generalisation, the OOV results are better when updating is not performed. 

\paragraph{\RQ[5] Overall, are some word embeddings better than others?}
Comparing the different word embedding techniques over our four sequence
labelling tasks, for the different evaluations (overall, out-of-domain
and OOV), there is no clear winner among the word embeddings -- for
\pos, \Skipgram appears to have a slight advantage, but this does not
generalise to other tasks.\\

While the aim of this paper was not to achieve the state of the art over
the respective tasks, it is important to concede that our best
(in-domain) results for \ner, \pos and \chunking are
slightly worse than the state of the art
(\tabref{benchmark}). The 2.7\% difference between our \ner system and
the best performing system is due to the fact that we use a first-order
instead of a second-order CRF~\cite{Ando:2005}, and for the other tasks,
there are similarly differences in the learner and the complexity of the
features used.
Another difference is that we tuned the hyperparameters with random
search, to enable replication using the same random seed.
In contrast, the hyperparameters for the state-of-the-art methods are
tuned more extensively by experts, making them more difficult to reproduce.

\section{Related Work}

\newcite{collobert2011natural} proposed a unified neural network framework
that learns word embeddings and applied it for \pos, \chunking, \ner and semantic role labelling. 
When they combined word embeddings with hand crafted features
(e.g., word suffixes for \pos;  gazetteers for \ner) and applied other
tricks like cascading and classifier combination, they achieved state-of-the-art performance.
Similarly, \newcite{turian2010word} evaluated three different word representations on \ner and \chunking, and concluded that unsupervised word representations improved \ner and \chunking. They also found that combining different word representations can further improve performance. \newcite{guo2014revisiting} also explored different ways of using word embeddings for \ner.  \newcite{owoputi2013improved} and \newcite{Schneider+:2014} found that \brown clustering enhances Twitter POS tagging and \mwe, respectively. Compared to previous work, we consider \textit{more} word representations including the most recent work and evaluate them on \textit{more} sequence labelling tasks, wherein the models are trained with training sets of varying size.

\newcite{Bansal+:2014} reported that direct use of word embeddings in dependency parsing did not show improvement. They achieved an improvement only when they performed hierarchical clustering of the word embeddings, and used features extracted from the cluster hierarchy.
In a similar vein, \newcite{Andreas:Klein:2014} explored the use of word embeddings for constituency parsing and concluded that the
information contained in word embeddings might duplicate the one acquired by a
syntactic parser, unless the training set is extremely small.
Other syntactic parsing studies that reported improvements by using
word embeddings include \newcite{Koo:2008}, \newcite{Koo:2010},
\newcite{Haffari:2011}, \newcite{Tratz:2011} and \newcite{chen:2014}.

Word embeddings have also been applied to other (non-sequential NLP)
tasks like grammar induction \cite{Spitkovsky:2011}, and semantic tasks
such as semantic relatedness, synonymy detection, concept
categorisation, selectional preference learning and analogy \cite{baroni:2014}.

\newcite{Huang:2009} demonstrated that using distributional word representations methods (like TF-IDF and LSA) as features, improves the labelling of OOV, when test for \pos and \chunking. In our study, we evaluate the labelling performance of OOV words for updated vs. not updated word embeddings representations, relative to the training set and with out-of-domain data.

\section{Conclusions}

We have performed an extensive extrinsic evaluation of four word embedding methods
under fixed experimental conditions, and evaluated their applicability to four sequence labelling tasks: \pos, \chunking, \ner and \mwe identification.
We found that word embedding features reliably outperformed unigram
features, especially with limited training data, but that there was
relatively little difference over Brown clusters, and
no one embedding method was consistently superior across the different tasks and settings.
Word embeddings and Brown clusters were also found to improve
out-of-domain performance and for OOV words.
We expected a performance gap between the fixed and task-updated embeddings, but the observed difference was marginal.
Indeed, we found that updating can result in overfitting.
We also carried out preliminary analysis of the impact of updating on
the vectors, a direction which we intend to pursue further.





\bibliographystyle{acl2013}
\bibliography{biblio}

\end{document}